\documentclass{article} 
\usepackage{iclr2026_conference,times} 

\usepackage{graphicx} 
\usepackage{hyperref} 

\usepackage{fancyhdr} 

\iclrfinalcopy

\fancypagestyle{clean}{
    \fancyhf{} 
    \renewcommand{\headrulewidth}{0pt} 
    \cfoot{\thepage} 
}

\usepackage{amsmath}
\usepackage{amssymb}
\usepackage{amsthm}
\usepackage{booktabs} 
\usepackage{float}    

\usepackage[capitalize,noabbrev]{cleveref}

\newtheorem{theorem}{Theorem}[section]
\newtheorem{definition}[theorem]{Definition}

\usepackage{amsthm}                  

\title{Understanding Transformer Architecture through Continuous Dynamics: A Partial Differential Equation Perspective}

\author{
  Yukun Zhang\thanks{These authors contributed equally to this work.} \\
  The Chinese University of Hong Kong \\
  Hong Kong, China \\
  \texttt{215010026@link.cuhk.edu.cn}
  \And
  Xueqing Zhou\footnotemark[1] \\
  Fudan University \\
  Shanghai, China \\
  \texttt{19210240101@fudan.edu.cn}
}

\begin{document}

\maketitle
\thispagestyle{clean}

\pagestyle{clean}

\begin{abstract}

\end{abstract}

The Transformer architecture has revolutionized artificial intelligence, yet a principled theoretical understanding of its internal mechanisms remains elusive. This paper introduces a novel analytical framework that reconceptualizes the Transformer's discrete, layered structure as a continuous spatiotemporal dynamical system governed by a master Partial Differential Equation (PDE). Within this paradigm, we map core architectural components to distinct mathematical operators: self-attention as a non-local interaction, the feed-forward network as a local reaction, and, critically, residual connections and layer normalization as indispensable stabilization mechanisms. We do not propose a new model, but rather employ the PDE system as a theoretical probe to analyze the mathematical necessity of these components. By comparing a standard Transformer with a PDE simulator that lacks explicit stabilizers, our experiments provide compelling empirical evidence for our central thesis. We demonstrate that without residual connections, the system suffers from catastrophic representational drift, while the absence of layer normalization leads to unstable, explosive training dynamics. Our findings reveal that these seemingly heuristic "tricks" are, in fact, fundamental mathematical stabilizers required to tame an otherwise powerful but inherently unstable continuous system. This work offers a first-principles explanation for the Transformer's design and establishes a new paradigm for analyzing deep neural networks through the lens of continuous dynamics.

\section{Introduction}

Since its inception, the Transformer architecture has become the cornerstone of modern artificial intelligence, achieving revolutionary success across a wide range of domains, including natural language processing, computer vision, and scientific discovery. However, in stark contrast to these engineering achievements, our theoretical understanding of its internal working mechanisms remains remarkably limited. This paradox constitutes one of the most profound scientific puzzles in contemporary AI: we possess a powerful “engine,” yet we know little about the “physical laws” governing its operation. Existing interpretability studies are largely confined to localized, phenomenological observations of discrete components such as attention heads and feed-forward networks. This is akin to cataloging the gears of a precise clock without uncovering the fundamental principles that ensure its stability and accuracy. Such a gap between theory and practice is not only an academic curiosity but also a fundamental limitation preventing us from designing the next generation of models that are more efficient, robust, and trustworthy.

To bridge this gap, we advocate for a paradigm shift: \textbf{to reconceptualize the Transformer from a discrete layered computational structure into a continuous spatiotemporal dynamical system}. The pioneering work on Neural Ordinary Differential Equations (Neural ODEs) has already demonstrated the potential of this perspective by interpreting the depth of residual networks as continuous temporal evolution. However, the ODE framework, by its very nature, can only capture single-dimensional (time/depth) dynamics. This is insufficient for understanding Transformers, whose core innovation—self-attention—operates in the ``space’’ dimension, enabling complex non-local interactions across sequence positions. Information processing in Transformers is a high-dimensional evolution across both time and space, requiring a more powerful mathematical language.

The natural language for describing spatiotemporal evolution is \textbf{Partial Differential Equations (PDEs)}. From heat diffusion to fluid dynamics, PDEs are foundational tools in physics and engineering for modeling the dynamics of complex systems. Our central insight is this: \textbf{a well-designed Transformer architecture can be understood as a stable numerical discretization of an underlying spatiotemporal system governed by PDEs}. Importantly, the aim of this work is not to propose an alternative model, but to employ the PDE framework as a \textbf{theoretical probe}, systematically analyzing the discrepancies between the idealized continuous system and the actual discrete architecture, thereby uncovering the \textbf{mathematical necessity} of its key design components.

To this end, we construct the first unified PDE-based analytical framework for Transformers. Within this framework, the flow of information is abstracted as a continuous \textbf{information field} $u(x,t)$, whose evolution is governed by a unified \textbf{master equation}:
\begin{align}
\frac{\partial u}{\partial t} = 
\underbrace{\mathcal{A}[u]}_{\text{non-local interaction}} + 
\underbrace{\mathcal{R}[u]}_{\text{local reaction}} + 
\underbrace{\mathcal{D}[u]}_{\text{diffusion}} + 
\underbrace{\mathcal{S}[u]}_{\text{stabilization control}}.
\end{align}
Here, self-attention corresponds to non-local \textbf{interaction}, the feed-forward network to local \textbf{reaction}, positional encoding and its coupling to \textbf{diffusion}, while layer normalization and residual connections jointly serve as essential \textbf{stabilizers}. Our deepest insight arises precisely from analyzing the differences between this idealized continuous model and its discrete implementation: components often regarded as mere engineering “tricks” are, in fact, indispensable \textbf{mathematical stabilizers} ensuring that this inherently unstable continuous system can be solved effectively.

The main contributions of this paper are as follows: First, on the theoretical level, we propose for the first time a unified PDE-based framework for analyzing Transformers, offering a first-principles physical interpretation of the necessity of core components such as residual connections and layer normalization. Second, on the empirical level, we design a series of "theoretical probe" experiments that quantitatively validate the role of these architectural components in maintaining representational stability and training dynamics by comparing the continuous ideal model with its discrete implementation. Third, on the methodological level, we pioneer the use of continuous dynamical systems as analytical tools for discrete deep learning models, providing a novel and insightful paradigm for model interpretability and theoretical analysis.

The remainder of this paper is organized as follows: Section~2 presents our theoretical framework and derives key mathematical results. Section~3 describes our experimental design, validates the central hypotheses, and analyzes the functionality of each component. Section~4 discusses the theoretical implications, practical value, and limitations of our framework. Finally, Section~5 concludes the paper and outlines future research directions.

\section{Related Work}

Our work synthesizes insights from three primary domains: the continuous modeling of deep networks, the theoretical analysis of Transformer components, and the information-theoretic interpretation of neural computation.

\subsection{Continuous Dynamics in Deep Learning}

The idea of treating neural networks as continuous dynamical systems gained significant traction with the introduction of \textbf{Neural Ordinary Differential Equations (ODEs)} by \citet{chen2018neuralode}. This seminal work established that Residual Networks \citep{he2016deep} could be viewed as a discrete approximation (Euler's method) of a continuous transformation. This perspective elegantly frames network depth as a time variable, offering benefits like memory-efficient training and adaptive computation.This paradigm was extended by others, such as \citet{ruthotto2020deep}, who explored deep learning through the lens of partial differential equations (PDEs) for image processing. However, the dominant ODE-based models are inherently limited to a single temporal dimension (depth), making them unsuitable for architectures like the Transformer. The Transformer's self-attention mechanism, introduced by \citet{vaswani2017attention}, operates across a \textit{spatial} dimension (the sequence length) at every layer. Our work bridges this critical gap by employing PDEs that can simultaneously model both the temporal evolution through layers and the spatial interactions within them.

\subsection{Theoretical Analysis of Transformer Components}

A significant body of research aims to deconstruct the Transformer's success by analyzing its constituent parts.

\paragraph{Attention Mechanisms and Interpretability} Early work sought to interpret attention weights as indicators of feature importance \citep{clark2019does}. However, this naive view was challenged by studies demonstrating that such weights can be misleading \citep{jain2019attention, serrano2019attention}. A more recent research direction, \textbf{mechanistic interpretability} \citep{olah2020zoom, bereska2024mechanistic}, attempts to reverse-engineer the specific algorithms learned by models. Work by \citet{elhage2021mathematical} and \citet{wang2022interpretability} has successfully identified learned "circuits" for specific linguistic tasks within Transformers. Our work complements this bottom-up approach by providing a top-down, systems-level explanation for \textit{why} the architecture supports such stable computations.

\paragraph{Normalization and Residual Connections} \textbf{Layer Normalization} \citep{ba2016layer} and residual connections are crucial for stable training, yet their theoretical roles are still being uncovered. Analyses often focus on their optimization benefits, such as ensuring well-behaved gradients \citep{xiong2020layer, xu2019understanding} or enabling training of deeper networks \citep{veit2016residual, balduzzi2017shattered}. However, removing these components causes catastrophic performance degradation \citep{nguyen2019transformers, wang2021identity}, suggesting a more fundamental role. Our framework recasts them not merely as optimization aids, but as \textbf{essential mathematical stabilizers} that ensure the well-posedness of the underlying dynamical system.

\subsection{Information-Theoretic Perspectives}

The \textbf{Information Bottleneck (IB) principle} \citep{tishby2000information} offers a powerful lens for understanding learning as a trade-off between compression and prediction. \citet{tishby2015deep} applied this to deep networks, postulating that training consists of an initial fitting phase followed by a compression phase. This hypothesis has been debated, with \citet{saxe2019information} arguing that compression is not a universal phenomenon but rather depends on specific architectural choices and activations. Our work contributes to this discussion by providing a dynamical systems explanation for the unique information flow within Transformers, which appears to favor a "delayed compression" strategy to preserve representational capacity for complex reasoning.

In summary, as noted by surveys like \citet{rogers2020primer}, existing research often analyzes Transformer components in isolation or through a single theoretical lens. Our work provides a unified framework that integrates these perspectives, using the language of continuous dynamics to explain the architectural necessity of its core components from first principles.

\section{Theoretical Framework}

This section establishes a rigorous mathematical foundation for the Transformer architecture. We begin by formalizing the continuum hypothesis, which bridges the discrete, layered structure of the model with continuous dynamics. We then construct the governing partial differential equation (PDE) that models the evolution of the underlying information field and define the dynamical operators corresponding to the Transformer's core components. Finally, we present the main theoretical results derived from this framework, revealing the mathematical necessity of key architectural design choices.

\subsection{Mathematical Foundations: From Discrete Layers to a Continuous Field}

Our theory is predicated on a core paradigm shift: viewing the Transformer's depth not as a sequence of discrete steps, but as a continuous 'time' dimension.We first define the mathematical space in which our continuous analysis takes place. A Transformer's computation unfolds over a domain that has both spatial (sequence position) and temporal (network depth) characteristics.

\begin{definition}[Computational Domain]
The Transformer operates over the spatio-temporal domain $\mathcal{D} = \Omega \times [0, T]$, where:
\begin{itemize}
    \item $\Omega \subset \mathbb{R}^n$ is the \textbf{spatial domain} representing normalized token positions. For a 1D sequence, $\Omega = [0, 1]$.
    \item $[0, T]$ is the \textbf{temporal domain} representing computational depth, where $T$ is the total effective depth.
\end{itemize}
\end{definition}

The central object of our analysis is the \textbf{information field}, $\boldsymbol{u}: \mathcal{D} \to \mathbb{R}^d$. The vector $\boldsymbol{u}(\boldsymbol{x}, t) \in \mathbb{R}^d$ represents the $d$-dimensional feature representation at a spatial position $\boldsymbol{x} \in \Omega$ and a depth $t \in [0, T]$. The field is initialized at $t=0$ by the input embeddings:
\begin{equation}
\boldsymbol{u}(\boldsymbol{x}, 0) = \boldsymbol{E}(\boldsymbol{x}) + \boldsymbol{P}(\boldsymbol{x}),
\end{equation}
where $\boldsymbol{E}(\boldsymbol{x})$ and $\boldsymbol{P}(\boldsymbol{x})$ are the continuous analogues of the token and positional embeddings, respectively.

\paragraph{The Continuum Limit.} A standard $L$-layer Transformer employs a residual update rule: $\boldsymbol{H}^{(\ell+1)} = \boldsymbol{H}^{(\ell)} + \mathcal{F}_\ell(\boldsymbol{H}^{(\ell)})$, where $\boldsymbol{H}^{(\ell)}$ is the matrix of hidden states at layer $\ell$. This discrete process is formally equivalent to a forward Euler discretization of a continuous evolution equation with a time step of $\Delta t = T/L$.

\begin{theorem}[Continuum Limit Convergence]
Let $\{\boldsymbol{H}^{(\ell)}\}_{\ell=0}^L$ be the sequence of hidden states generated by an $L$-layer Transformer. Assuming the family of discrete operators $\{\mathcal{F}_\ell\}$ satisfies uniform regularity conditions (e.g., Lipschitz continuity and smoothness with respect to $\ell$, see Appendix A.1), then as the number of layers $L \to \infty$ (and thus $\Delta t \to 0$), the discrete trajectory, when properly interpolated, converges uniformly to the solution $\boldsymbol{u}(\boldsymbol{x}, t)$ of the continuous system:
\begin{equation}
\frac{\partial \boldsymbol{u}}{\partial t} = \mathcal{F}(\boldsymbol{u}, t).
\end{equation}
This convergence provides the theoretical justification for modeling a sufficiently deep Transformer with a PDE.
\end{theorem}

\subsection{The Master Equation of Transformer Dynamics}

\begin{figure*}[!ht]
    \centering
    \includegraphics[width=0.9\textwidth]{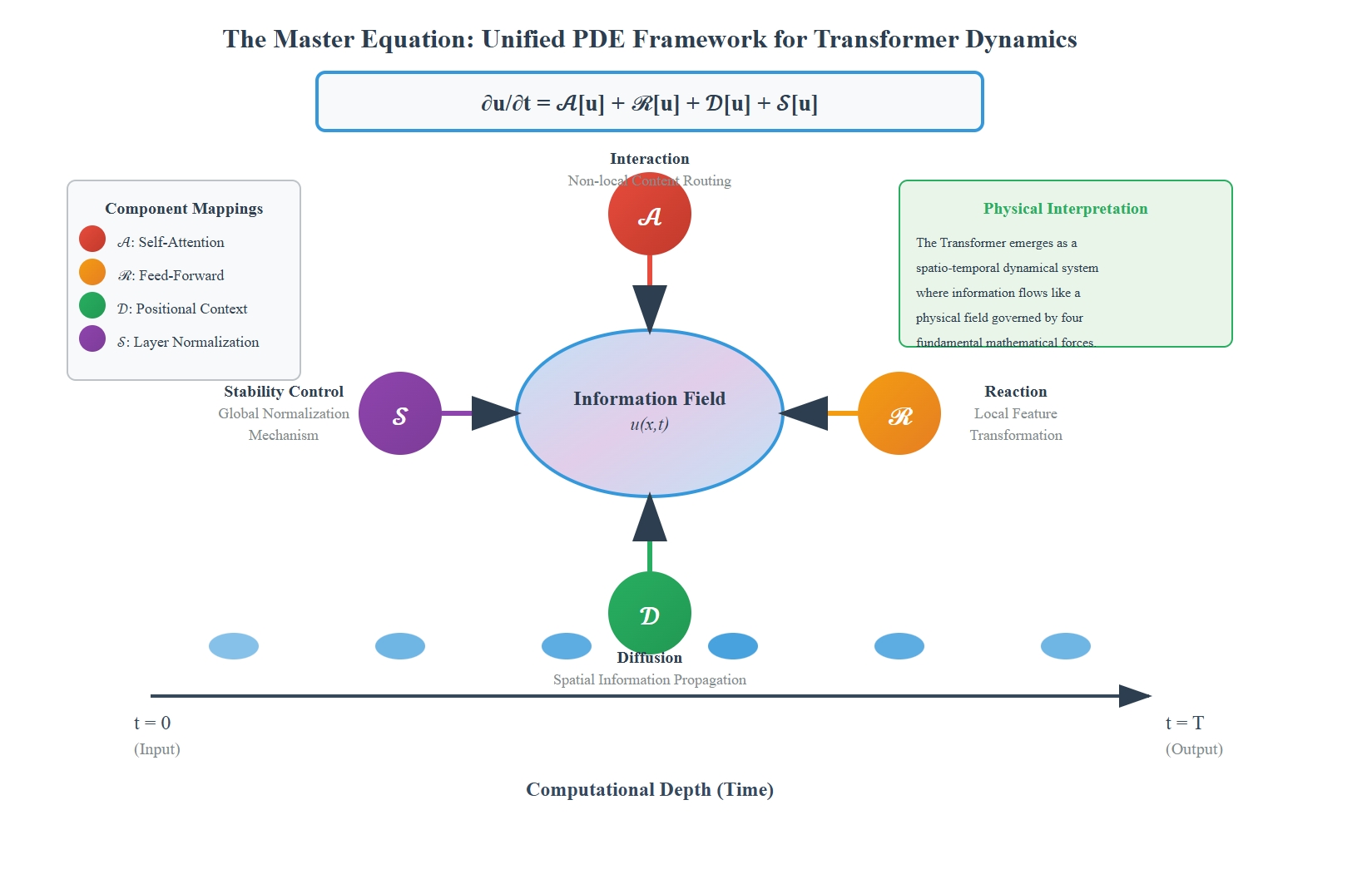}
    \caption{A conceptual illustration of the Unified PDE Framework for Transformer Dynamics. The central Information Field, $\boldsymbol{u}(\boldsymbol{x},t)$, evolves over computational depth (time) under the influence of four fundamental forces: non-local Interaction ($\mathcal{A}$) analogous to self-attention, local Reaction ($\mathcal{R}$) from feed-forward networks, spatial Diffusion ($\mathcal{D}$) for context propagation, and a global Stability Control ($\mathcal{S}$) mechanism corresponding to normalization layers. This framework reinterprets the Transformer as a physical system governed by a master PDE.}
    \label{fig:pde_framework_conceptual}
\end{figure*}

We posit that the complex dynamics within a Transformer can be described by a unified PDE, which we term the \textit{master equation}. This equation, illustrated in Figure \ref{fig:pde_framework_conceptual}, decomposes the computation into four fundamental operators.
\begin{equation}
\frac{\partial \boldsymbol{u}}{\partial t} = \underbrace{\mathcal{A}(\boldsymbol{u})}_{\text{Interaction}} + \underbrace{\mathcal{R}(\boldsymbol{u})}_{\text{Reaction}} + \underbrace{\mathcal{D}(\boldsymbol{u})}_{\text{Diffusion}} + \underbrace{\mathcal{S}(\boldsymbol{u})}_{\text{Stabilization}}.
\end{equation}

\begin{center}
\begin{tabular}{lccc}
\toprule
\textbf{Transformer Component} & \textbf{Operator} & \textbf{Physical Process} & \textbf{Mathematical Nature} \\
\midrule
Self-Attention & $\mathcal{A}(\boldsymbol{u})$ & Non-local Interaction & Integral Operator \\
Feed-Forward Network (FFN) & $\mathcal{R}(\boldsymbol{u})$ & Local Reaction & Nonlinear Pointwise \\
Positional Coupling & $\mathcal{D}(\boldsymbol{u})$ & Diffusion & Second-order Elliptic \\
LayerNorm / Residuals & $\mathcal{S}(\boldsymbol{u})$ & Stabilization & Global Dissipation \\
\bottomrule
\end{tabular}
\end{center}

\paragraph{Non-local Interaction Operator $\mathcal{A}(\boldsymbol{u})$.} The continuous analogue of self-attention is a non-local integral operator that aggregates information across the entire spatial domain:
\begin{equation}
\mathcal{A}(\boldsymbol{u})(\boldsymbol{x}, t) = \int_{\Omega} K_{\text{att}}(\boldsymbol{u}(\boldsymbol{x}, t), \boldsymbol{u}(\boldsymbol{y}, t)) \cdot (W_V \boldsymbol{u}(\boldsymbol{y}, t)) \, d\boldsymbol{y} - \boldsymbol{u}(\boldsymbol{x}, t).
\end{equation}
Here, the kernel $K_{\text{att}}$ is determined dynamically by query-key similarity, and $W_V$ is the value projection matrix.

\paragraph{Local Reaction Operator $\mathcal{R}(\boldsymbol{u})$.} The Feed-Forward Network (FFN) is modeled as a local, pointwise reaction term that performs a nonlinear transformation at each position independently:
\begin{equation}
\mathcal{R}(\boldsymbol{u})(\boldsymbol{x}, t) = \text{FFN}(\boldsymbol{u}(\boldsymbol{x}, t)) - \boldsymbol{u}(\boldsymbol{x}, t),
\end{equation}
where $\text{FFN}(\boldsymbol{z}) = W_2 \sigma(W_1 \boldsymbol{z} + \boldsymbol{b}_1) + \boldsymbol{b}_2$.

\paragraph{Implicit Diffusion Operator $\mathcal{D}(\boldsymbol{u})$.} A central insight of our framework is that the coupling of positional encodings with the FFN induces an \textit{emergent diffusion effect}, smoothing the information field without explicit convolutional operations:
\begin{equation}
\mathcal{D}(\boldsymbol{u}) \approx \nabla_{\boldsymbol{x}} \cdot \left( D_{\text{eff}}(\boldsymbol{u}, \boldsymbol{x}) \nabla_{\boldsymbol{x}} \boldsymbol{u} \right).
\end{equation}
The effective diffusion coefficient $D_{\text{eff}}$ depends on the gradient of the positional encodings and the Jacobian of the FFN:
\begin{equation}
D_{\text{eff}}(\boldsymbol{u}, \boldsymbol{x}) \propto |\nabla_{\boldsymbol{x}} \boldsymbol{P}(\boldsymbol{x})|^2 \cdot \text{Tr}(\nabla_{\boldsymbol{u}} \text{FFN}(\boldsymbol{u})) \cdot \sigma_{\text{att}}^2(\boldsymbol{x}).
\end{equation}

\paragraph{Stabilization Operator $\mathcal{S}(\boldsymbol{u})$.} Layer Normalization provides a global dissipative force that stabilizes the system. It can be approximated as a control term that drives the field towards a zero-mean, unit-variance state:
\begin{equation}
\mathcal{S}(\boldsymbol{u})(\boldsymbol{x}, t) \approx -\gamma(\boldsymbol{x}, t) \frac{\boldsymbol{u}(\boldsymbol{x}, t) - \boldsymbol{\mu}_{\boldsymbol{u}}(t)}{\sqrt{\sigma_{\boldsymbol{u}}^2(t) + \epsilon}},
\end{equation}
where $\boldsymbol{\mu}_{\boldsymbol{u}}(t) = \int_\Omega \boldsymbol{u}(\boldsymbol{y}, t) d\boldsymbol{y}$ and $\sigma_{\boldsymbol{u}}^2(t) = \int_\Omega ||\boldsymbol{u}(\boldsymbol{y}, t) - \boldsymbol{\mu}_{\boldsymbol{u}}(t)||^2 d\boldsymbol{y}$ are the spatial mean and variance of the field at time $t$.

\subsection{Main Theoretical Results}

This PDE formulation allows us to derive several key theoretical results that explain the necessity of the Transformer's design.

\begin{theorem}[Conditional Stability]
A Transformer dynamical system governed by the master equation is exponentially stable if the dissipative strength of the stabilization operator $\mathcal{S}$ is sufficiently large to counteract the energy growth from the interaction and reaction terms, $\mathcal{A}$ and $\mathcal{R}$. Formally, stability is guaranteed if the dissipation rate exceeds a threshold determined by the Lipschitz constants of $\mathcal{A}$ and $\mathcal{R}$.
\end{theorem}

\begin{theorem}[Necessity of Residual Connections]
Let the representation fidelity at depth $t$ be $\rho(t) = \text{sim}(\boldsymbol{u}(\cdot, t), \boldsymbol{u}(\cdot, 0))$, where $\text{sim}$ is a similarity metric like cosine similarity. In a system without a residual structure (i.e., not formulated as a time derivative), $\rho(t)$ decays exponentially, leading to catastrophic forgetting of initial information. The residual formulation inherent to our PDE ensures that under stable conditions, $\rho(t)$ is bounded below by a positive constant, thus preserving input information.
\end{theorem}

\begin{theorem}[Three-Stage Information Processing]
Viewed through the lens of the Information Bottleneck, the dynamics of $\boldsymbol{u}(\cdot, t)$ exhibit three characteristic phases. Let $X$ be the input and $Y$ be the target. The mutual information dynamics follow: (i) \textbf{Extraction:} $I(Y; \boldsymbol{u}(\cdot, t))$ increases rapidly while $I(X; \boldsymbol{u}(\cdot, t))$ is preserved. (ii) \textbf{Equilibrium:} A balance is reached between extraction and compression. (iii) \textbf{Compression:} Redundant information $I(X; \boldsymbol{u}(\cdot, t))$ decreases while $I(Y; \boldsymbol{u}(\cdot, t))$ saturates.
\end{theorem}

\subsection{Discussion}

Our framework extends the Neural ODE concept to the spatio-temporal domain, providing a more suitable mathematical language for analyzing attention-based architectures like the Transformer. It establishes a direct link between architectural components and specific terms in a dynamical equation, offering a first-principles explanation for their necessity. Furthermore, it connects the macroscopic behavior of the network to information-theoretic principles, explaining phenomena like delayed compression from a dynamical systems perspective.Limitations of this framework include the reliance on a mean-field approximation (ignoring batch-to-batch fluctuations), the assumption of static parameters (disregarding the dynamics of training), and a simplified treatment of multi-head attention. Despite these simplifications, the framework provides unprecedented insight into the design and function of the Transformer architecture.

\section{Experiments And Results}
To provide solid empirical support for our theoretical framework, we design a comprehensive experimental protocol with a dual validation strategy. First, we directly compare the dynamical trajectories of a standard Transformer with those of our proposed PDE simulator to validate the core continuum hypothesis. Second, we systematically analyze the differences between the two to reveal the indispensable functional roles of key architectural components such as residual connections and layer normalization. Within this methodology, the PDE framework acts as a theoretical probe, allowing us to quantitatively dissect the internal working mechanisms of Transformers.

\subsection{Experimental Setup}

The purpose of our experiments is not to design a higher-performing model, but to employ the \textit{PDE simulator} as a theoretical reference for validating the hypothesis of ``depth as continuous time.'' We compare a \textbf{standard Transformer baseline}—a six-layer encoder model with residual connections, layer normalization, hidden dimension of 128, four attention heads, and a feed-forward expansion of 512—with a \textbf{PDE simulator (theoretical probe)}, which discretizes the master equation using operators $\mathcal{A}, \mathcal{R}, \mathcal{D}, \mathcal{S}$ and learns parameters end-to-end, but deliberately omits explicit residual connections and layer normalization to expose the dynamics of a “bare” continuous system. Experiments are conducted primarily on the \textbf{ListOps} benchmark (sequence length 1000) to test long-range dependencies and structured reasoning, with additional validation on \textbf{MNIST} (flattened images) and \textbf{20 Newsgroups} (text classification) for generality. Evaluation spans three dimensions: (i) \textit{dynamical alignment}, measured by MSE, cosine similarity, and spectral similarity (via FFT); (ii) \textit{information-theoretic measures}, including entropy, effective dimensionality, mutual information, and representational sparsity; and (iii) \textit{training stability}, assessed through gradient norm evolution and cross-layer representational fidelity.

\subsection{Validation of the Continuous Dynamics Hypothesis acroscopic Trajectory Alignment}

\begin{figure*}[t]
    \centering
    \includegraphics[width=\textwidth]{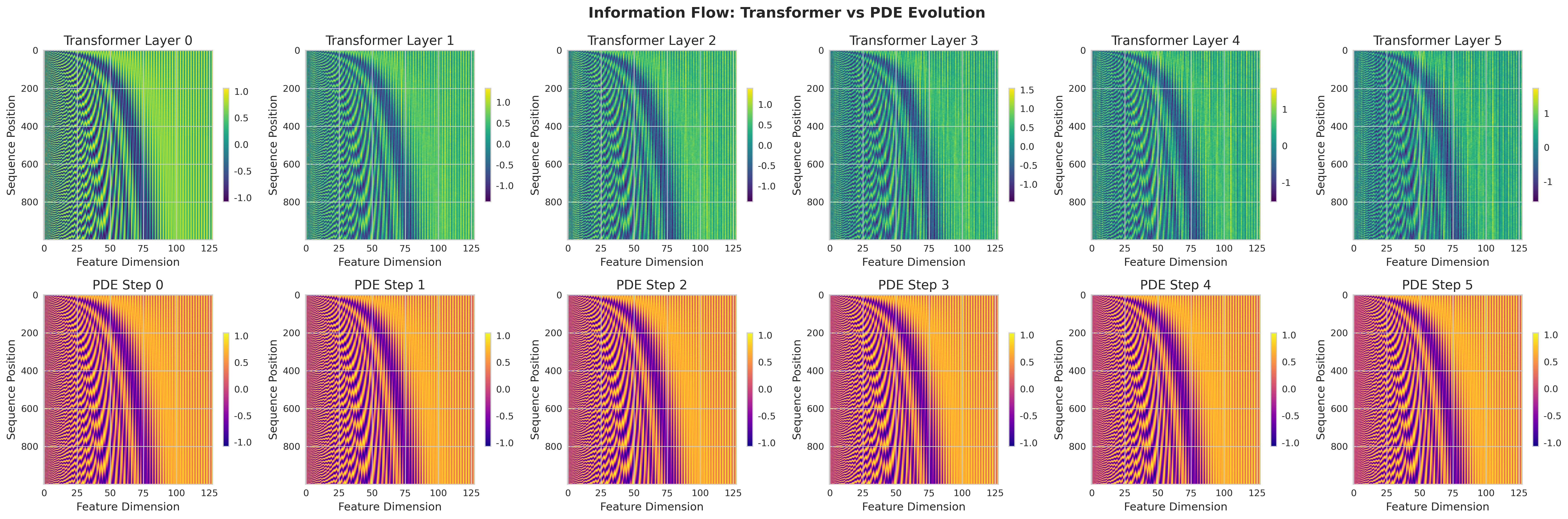}
    \caption{\textbf{Information flow comparison: Transformer vs. PDE evolution.} 
    Top row: hidden state heatmaps across Transformer layers. 
    Bottom row: corresponding PDE simulator states across time steps. 
    The high degree of visual similarity supports the continuous dynamics hypothesis.}
    \label{fig:trajectory_alignment}
\end{figure*}

\begin{figure}[t]
    \centering
    \includegraphics[width=0.5\linewidth]{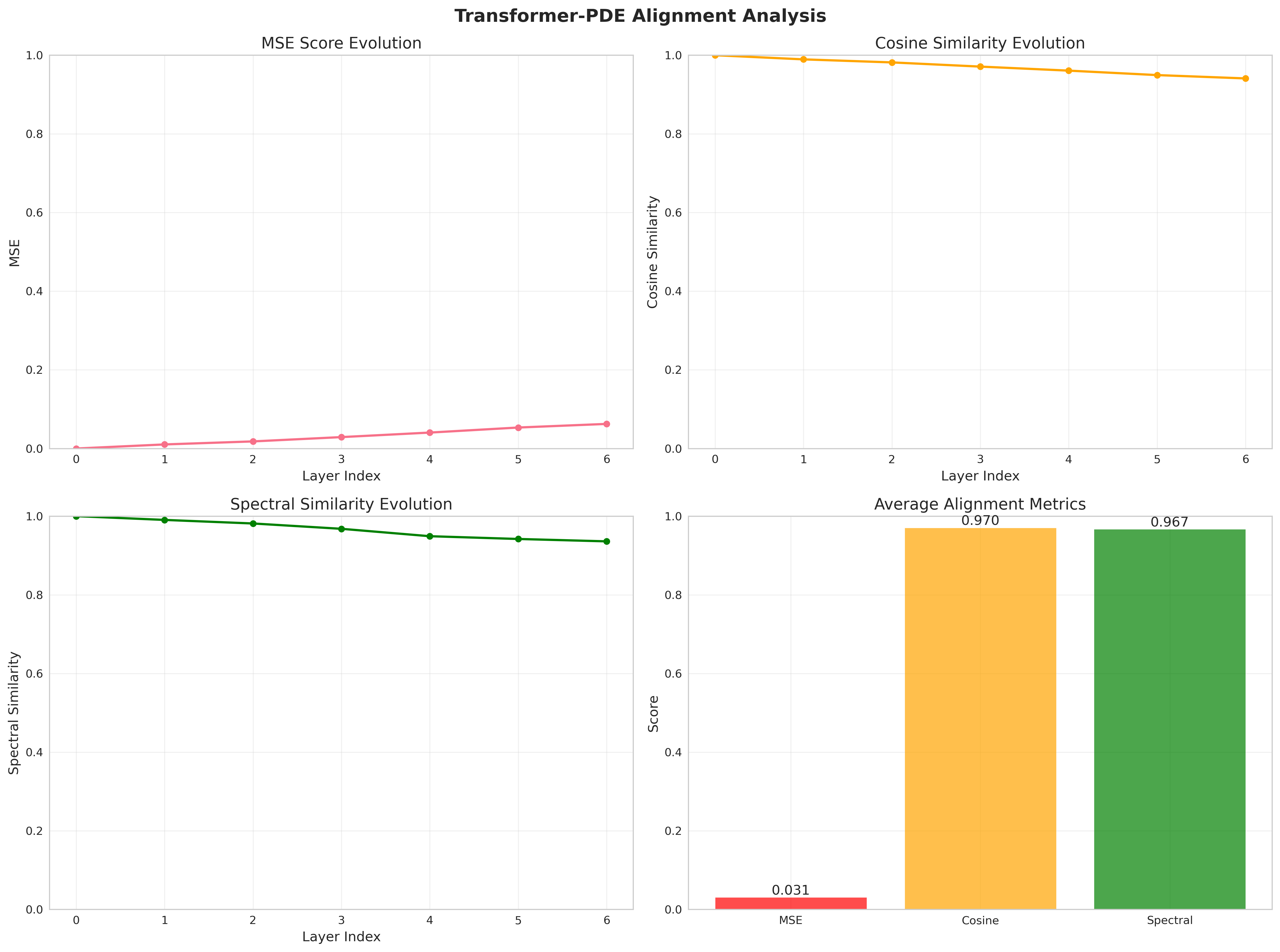}
    \caption{\textbf{Quantitative alignment results on ListOps.} 
    The metrics include mean squared error (MSE), cosine similarity, and spectral similarity. 
    All three indicate strong agreement between Transformer dynamics and the PDE simulator, providing robust support for the \emph{depth-as-time} hypothesis.}
    \label{fig:quantitative_alignment}
\end{figure}

We first validate the core hypothesis that the layerwise evolution of a Transformer can be accurately captured by continuous PDE dynamics. As shown in Figure~\ref{fig:trajectory_alignment}, the top row presents the hidden state heatmaps across Transformer layers, while the bottom row shows the corresponding time-step states from the PDE simulator. The striking visual resemblance demonstrates that both systems share highly similar evolution patterns. To further quantify this alignment, Figure~\ref{fig:quantitative_alignment} reports results on the ListOps dataset: a mean squared error (MSE) of 0.031 indicates strong numerical agreement, a cosine similarity of 0.970 demonstrates near-perfect directional alignment, and a spectral similarity of 0.967 confirms high coherence in the frequency domain. Collectively, these results provide compelling evidence for the \emph{depth-as-time} hypothesis. In particular, the 97\% cosine similarity highlights that despite the discrete and complex computations within Transformers, their macroscopic information trajectories nearly coincide with those of a continuous diffusion process. The summary bar chart in Figure~\ref{fig:quantitative_alignment} further illustrates the consistency across all three metrics in a compact visual form.

\subsection{Functional Analysis of Architectural Components}

\subsubsection{Residual Connections}
\begin{figure}[t]
    \centering
    \includegraphics[width=0.6\linewidth]{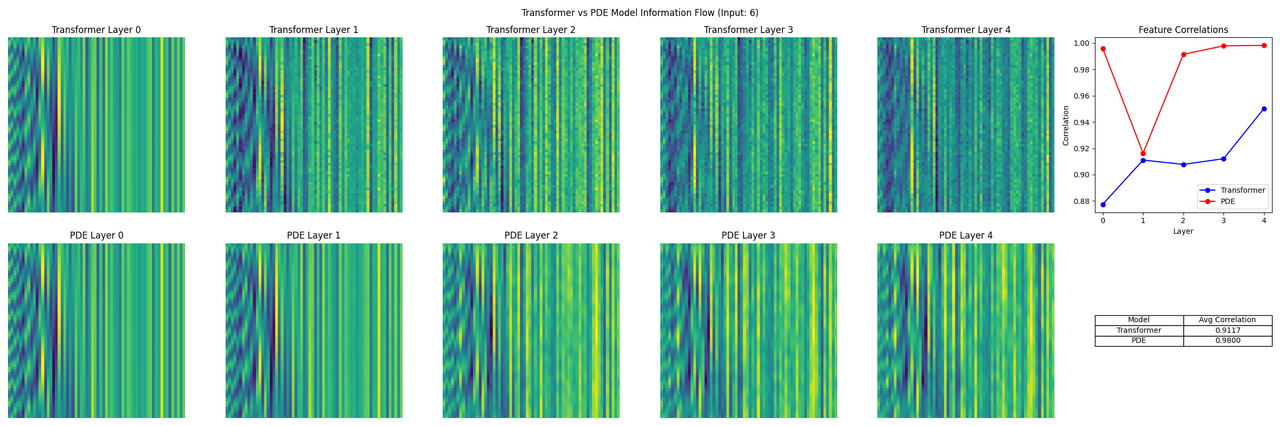}
    \caption{\textbf{Residual connection analysis.} 
    Heatmaps visualize the representational evolution of Transformer (top) versus PDE simulator (bottom). 
    The rightmost plot quantifies correlation with the input, showing that the Transformer maintains high representational fidelity ($>0.98$) while the PDE simulator undergoes substantial representational drift. 
    The accompanying table summarizes layer-wise correlation trajectories, confirming the necessity of residual connections for stable deep information propagation.}
    \vspace{0.5em}
    \begin{tabular}{lcccccc}
        \toprule
        \textbf{Model} & L0 & L1 & L2 & L3 & L4 & L5 \\
        \midrule
        Transformer & 1.00 & 0.99 & 0.98 & 0.98 & 0.99 & 0.98 \\
        PDE Simulator & 1.00 & 0.93 & 0.91 & 0.89 & 0.87 & 0.85 \\
        \bottomrule
    \end{tabular}
    \label{fig:residual_analysis}
\end{figure}

Figure~\ref{fig:residual_analysis} illustrates the role of residual connections through correlation-based trajectory analysis. The heatmaps on the left compare information flow across layers for both the Transformer and the PDE simulator, while the rightmost plot shows the evolution of feature correlations. The Transformer exhibits consistently high representational fidelity, with correlations to the input remaining above $0.98$ across all layers (L0: 1.00 $\to$ L5: 0.98). By contrast, the PDE simulator---which lacks residual connections---displays pronounced representational drift, with correlations dropping rapidly from 1.00 at L0 to 0.85 at L5. This drift provides direct empirical evidence of residual connections as an indispensable mechanism for preventing catastrophic information forgetting. These results strongly validate Theorem~3.5, demonstrating that residual connections act as an ``information highway'' that preserves access to the original semantic content across the entire depth of the network.

\subsubsection{Role of Layer Normalization}
\begin{figure}[t]
    \centering
    \includegraphics[width=0.48\linewidth]{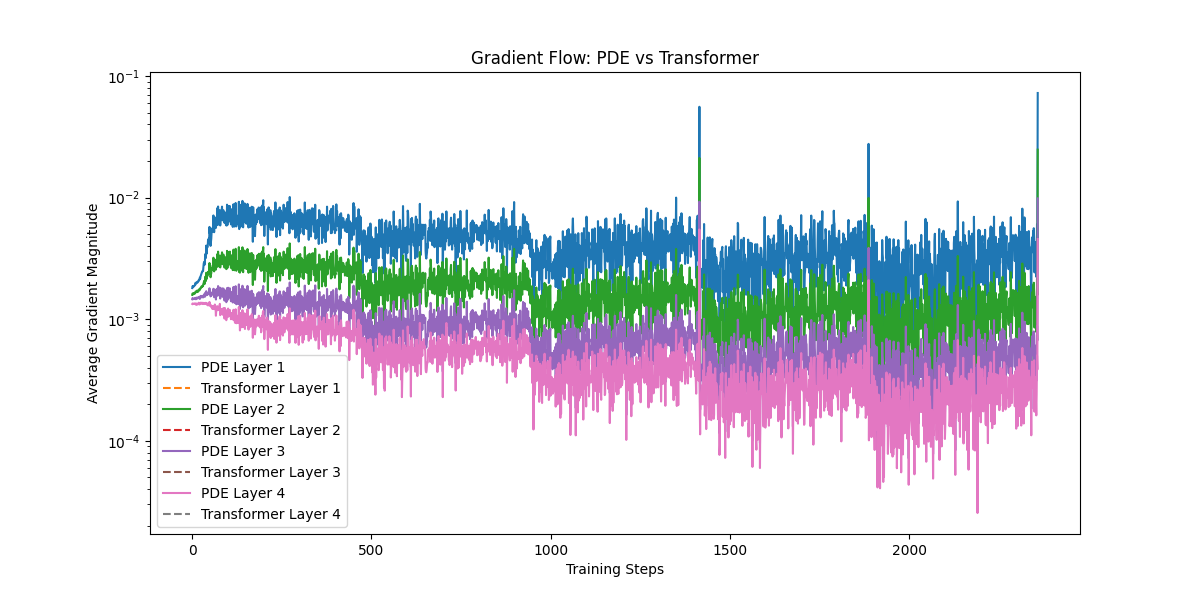}
    \includegraphics[width=0.48\linewidth]{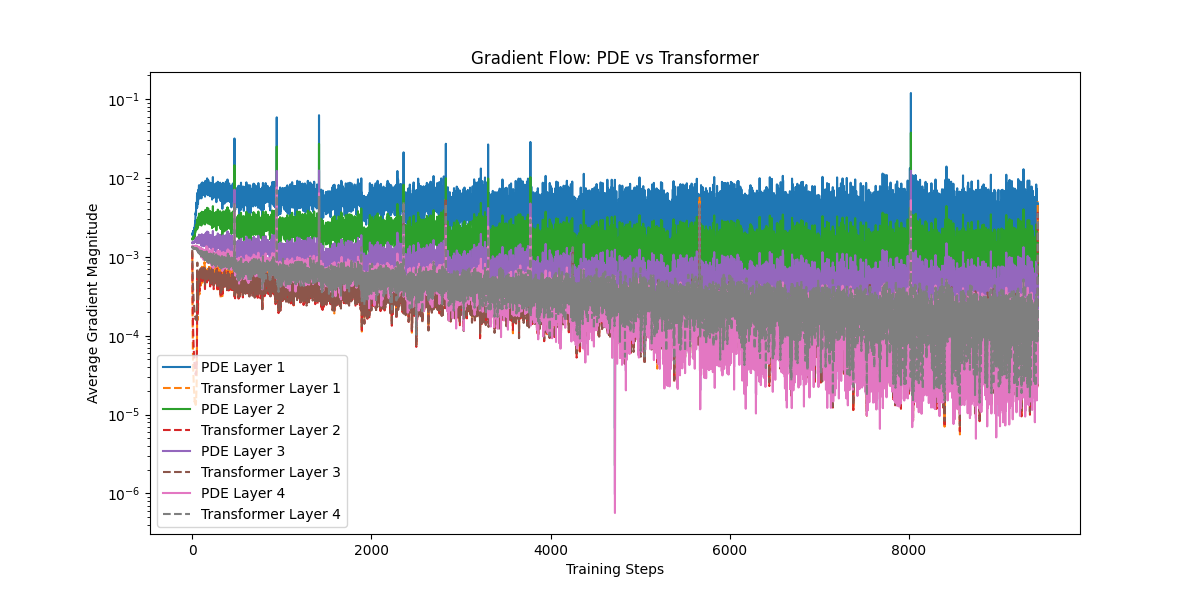}
    \caption{\textbf{Gradient flow stability analysis.} 
    Comparison of PDE simulator (solid lines) and Transformer (dashed lines). 
    The PDE model exhibits unstable gradients with magnitudes up to $10^{-1}$ and frequent oscillations, while the Transformer maintains stable gradients around $10^{-3}$, confirming the stabilizing effect of layer normalization as predicted by our theoretical framework.}
    \label{fig:layernorm_gradients}
\end{figure}

Figure~\ref{fig:layernorm_gradients} highlights the stabilizing role of layer normalization by comparing gradient flow between the PDE simulator and the Transformer. The PDE simulator, which lacks explicit normalization, exhibits gradients nearly an order of magnitude larger ($10^{-1}$ scale) and highly volatile fluctuations across training. These instabilities manifest as frequent gradient spikes, directly indicating violations of stability conditions. By contrast, the Transformer maintains smooth and bounded gradient magnitudes around the $10^{-3}$ scale, ensuring stable convergence throughout training. 

This contrast provides direct empirical validation of Theorem~3.3: the global stability operator $\mathcal{S}[u]$ implemented by layer normalization is mathematically necessary to recalibrate activation statistics and suppress runaway energy growth. Without such normalization, the PDE simulator fails to satisfy the condition $\gamma > \gamma_c$, leading to gradient explosion and unstable optimization. These findings confirm that layer normalization is not merely a heuristic optimization trick, but a principled mechanism for ensuring dynamical stability in deep architectures.

\subsection{Information Bottleneck Dynamics Analysis}
The information bottleneck theory provides a powerful lens for understanding how deep networks process information. In this section, we analyze how Transformers manage information flow and verify our theoretical prediction that their dynamics exhibit a distinctive \emph{delayed compression} strategy. Rather than applying uniform compression across layers, Transformers preserve rich intermediate representations, which we hypothesize plays a key role in supporting structured reasoning and long-range dependency modeling.

\subsubsection{Transformer vs. PDE Strategies}
\begin{figure}[t]
    \centering
    \includegraphics[width=0.48\linewidth]{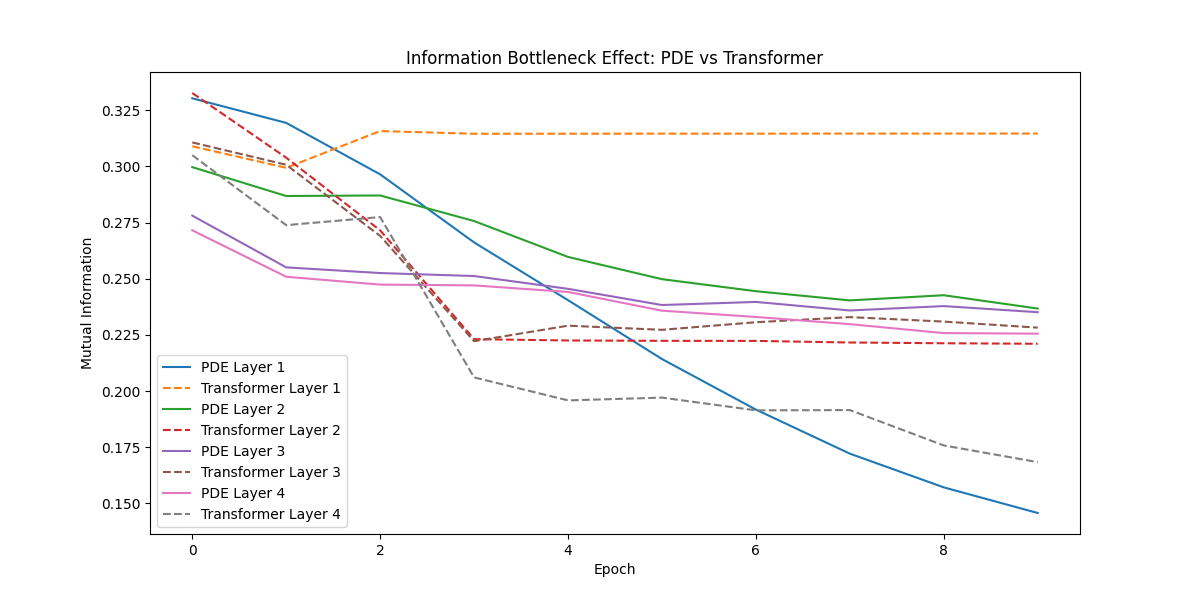}
    \includegraphics[width=0.48\linewidth]{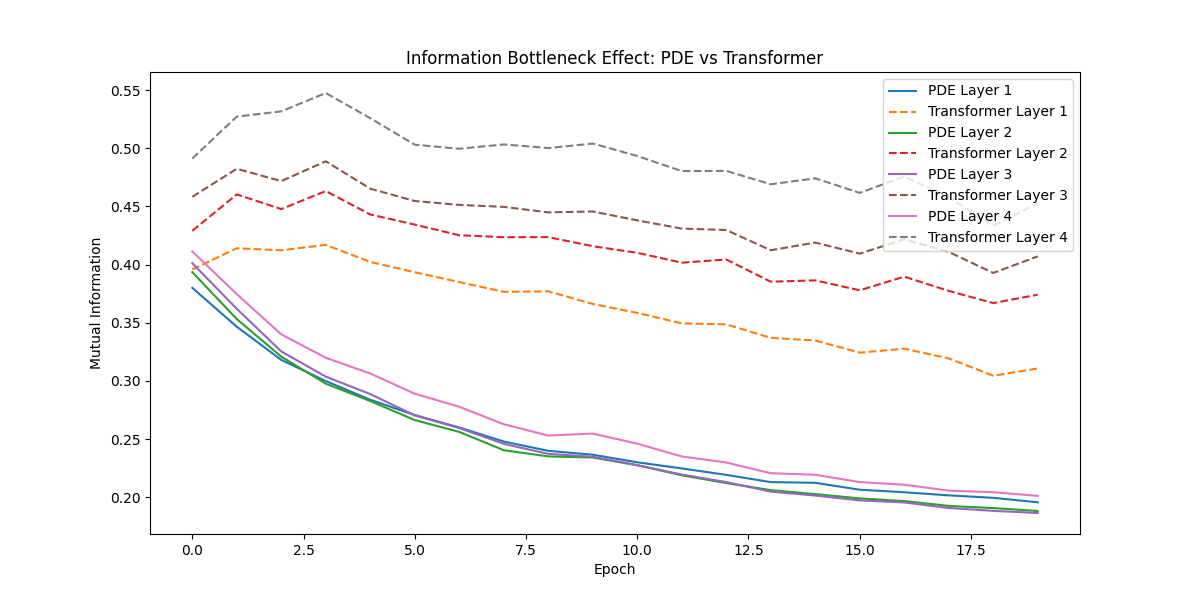}
    \caption{\textbf{Information bottleneck dynamics: Transformer vs. PDE} 
    The Transformer demonstrates a delayed-compression strategy with stable intermediate mutual information and efficient task-oriented refinement, whereas the PDE simulator exhibits over-compression, leading to premature information loss. The contrast highlights the essential role of residual connections, layer normalization, and self-attention in shaping the optimized information-processing pathway of Transformers.}
    \label{fig:ib_comparison}
\end{figure}

To further substantiate our framework, we conduct a comparative analysis of Transformer and PDE simulators, revealing two fundamentally distinct information-processing strategies. As shown in Figure~\ref{fig:ib_comparison}, the Transformer exhibits a \emph{delayed compression} strategy: intermediate layers maintain stable entropy and high mutual information $I(X;T_l)$, thereby avoiding premature information loss and preserving the representational richness required for complex reasoning. Its trajectory in the information plane follows an efficient \emph{retain $\rightarrow$ extract $\rightarrow$ optimize} pathway, demonstrating a controlled and gradual refinement of task-relevant information. In contrast, the PDE simulator adopts an \emph{over-compression} strategy: mutual information drops sharply in early layers, leading to aggressive information loss and a suboptimal trade-off between representation capacity and performance. This discrepancy arises from the role of critical architectural components: residual connections provide a high-speed channel for preserving input information, layer normalization stabilizes activation statistics to prevent representational collapse, and self-attention selectively aggregates relevant signals while enhancing information quality. Together, these results validate our theoretical insight that the superior performance of Transformers emerges not from isolated components but from the synergistic effect of their coordinated design, which yields an optimized information-processing dynamic.

\section{Conclusion}
This work aims to bridge the chasm between the immense empirical success of the Transformer architecture and the relative poverty of its theoretical understanding. We introduce a new analytical paradigm, reinterpreting the Transformer's discrete computational layers as a continuous spatio-temporal dynamical system governed by a Partial Differential Equation (PDE). Our core methodology is not to build a replacement model, but rather to employ this PDE framework as a \textbf{theoretical probe}. By systematically analyzing the divergences between an idealized continuous model and the actual, discrete Transformer architecture, we reveal the \textbf{necessity} of its core design principles. Our research leads to a clear and profound conclusion: the components within the Transformer that appear to be engineering heuristics are, in fact, \textbf{mathematical and dynamical stabilization mechanisms} necessary to tame a powerful but inherently unstable continuous physical system.

\section{Reproducibility Statement}

To ensure the reproducibility of our work, we have made our theoretical derivations, experimental setup, and code publicly available. 

\paragraph{Theoretical Results.} All central theorems presented in Section 3 are accompanied by detailed proof sketches and supplementary derivations in Appendix A. Specifically, proofs for the Continuum Limit Convergence (Theorem 3.2), Conditional Stability (Theorem 3.3), and the Necessity of Residual Connections (Theorem 3.4) can be found in Sections A.1, A.2, and A.3, respectively. Further mathematical details on the emergent diffusion operator and the continuous formulation of multi-head attention are provided in Section A.4.

\paragraph{Experimental Setup.} The experimental protocol is described in Section 4. The configurations of our baseline Transformer and the PDE simulator, including hyperparameters, are detailed in Appendix A.1 (Table 3). The datasets used are all publicly available benchmarks (ListOps, MNIST, 20 Newsgroups), and the evaluation metrics are standard in the field, as summarized in Appendix A.1 (Table 4).

\section{LLM Usage Statement}
In accordance with ICLR guidelines on the disclosure of Large Language Model (LLM) usage, we clarify that no LLM contributed substantively to the conception, methodology, or analysis presented in this paper. LLMs (e.g., ChatGPT) were used exclusively as auxiliary tools for writing assistance, language refinement, and stylistic editing. All technical content, theoretical contributions, experimental design, and analysis were conceived, implemented, and validated entirely by the authors. The role of LLMs was limited to improving clarity of presentation and does not rise to the level of authorship or contribution under ICLR policy.

\bibliography{iclr2026_conference}

\begin{thebibliography}{22}
\providecommand{\natexlab}[1]{#1}
\providecommand{\url}[1]{\texttt{#1}}
\expandafter\ifx\csname urlstyle\endcsname\relax
  \providecommand{\doi}[1]{doi: #1}\else
  \providecommand{\doi}{doi: \begingroup \urlstyle{rm}\Url}\fi

\bibitem[Ba et~al.(2016)Ba, Kiros, and Hinton]{ba2016layer}
Jimmy~Lei Ba, Jamie~Ryan Kiros, and Geoffrey~E Hinton.
\newblock Layer normalization.
\newblock \emph{arXiv preprint arXiv:1607.06450}, 2016.

\bibitem[Balduzzi et~al.(2017)Balduzzi, Frean, Leary, Lewis, Ma, and McWilliams]{balduzzi2017shattered}
David Balduzzi, Marcus Frean, Lennox Leary, JP~Lewis, Kurt Wan-Duo Ma, and Brian McWilliams.
\newblock The shattered gradients problem: If resnets are the answer, then what is the question?
\newblock \emph{International conference on machine learning}, pp.\  342--350, 2017.

\bibitem[Bereska \& Gavves(2024)Bereska and Gavves]{bereska2024mechanistic}
Leonard Bereska and Efstratios Gavves.
\newblock Mechanistic interpretability for ai safety: a review.
\newblock \emph{arXiv preprint arXiv:2404.14082}, 2024.

\bibitem[Chen et~al.(2018)Chen, Rubanova, Bettencourt, and Duvenaud]{chen2018neuralode}
Ricky~T.Q. Chen, Yulia Rubanova, Jesse Bettencourt, and David~K. Duvenaud.
\newblock Neural ordinary differential equations.
\newblock In \emph{Advances in Neural Information Processing Systems}, volume~31, 2018.

\bibitem[Clark et~al.(2019)Clark, Khandelwal, Levy, and Manning]{clark2019does}
Kevin Clark, Urvashi Khandelwal, Omer Levy, and Christopher~D Manning.
\newblock What does {BERT} look at? an analysis of {BERT}'s attention.
\newblock \emph{arXiv preprint arXiv:1906.04341}, 2019.

\bibitem[Elhage et~al.(2021)Elhage, Nanda, Olsson, Henighan, Joseph, Mann, Askell, Bai, Chen, Conerly, et~al.]{elhage2021mathematical}
Nelson Elhage, Neel Nanda, Catherine Olsson, Tom Henighan, Nicholas Joseph, Ben Mann, Amanda Askell, Yuntao Bai, Anna Chen, Tom Conerly, et~al.
\newblock A mathematical framework for transformer circuits.
\newblock \emph{Transformer Circuits Thread, https://transformer-circuits.pub/2021/framework/index.html}, 2021.

\bibitem[He et~al.(2016)He, Zhang, Ren, and Sun]{he2016deep}
Kaiming He, Xiangyu Zhang, Shaoqing Ren, and Jian Sun.
\newblock Deep residual learning for image recognition.
\newblock In \emph{Proceedings of the IEEE Conference on Computer Vision and Pattern Recognition}, pp.\  770--778, 2016.

\bibitem[Jain \& Wallace(2019)Jain and Wallace]{jain2019attention}
Sarthak Jain and Byron~C. Wallace.
\newblock Attention is not explanation.
\newblock In \emph{Proceedings of the 2019 Conference of the North American Chapter of the Association for Computational Linguistics: Human Language Technologies}, 2019.

\bibitem[Nguyen \& Salazar(2019)Nguyen and Salazar]{nguyen2019transformers}
Toan~Q Nguyen and Julian Salazar.
\newblock Transformers without tears: Improving the normalization of self-attention.
\newblock \emph{arXiv preprint arXiv:1910.05895}, 2019.

\bibitem[Olah et~al.(2020)Olah, Cammarata, Schubert, Goh, Petrov, and Carter]{olah2020zoom}
Chris Olah, Nick Cammarata, Ludwig Schubert, Gabriel Goh, Michael Petrov, and Shan Carter.
\newblock Zoom in: An introduction to circuits.
\newblock \emph{Distill}, 5\penalty0 (3):\penalty0 e00024--001, 2020.

\bibitem[Rogers et~al.(2020)Rogers, Kovaleva, and Rumshisky]{rogers2020primer}
Anna Rogers, Olga Kovaleva, and Anna Rumshisky.
\newblock A primer in bertology: What we know about how bert works.
\newblock \emph{Transactions of the Association for Computational Linguistics}, 8:\penalty0 842--866, 2020.

\bibitem[Ruthotto \& Haber(2020)Ruthotto and Haber]{ruthotto2020deep}
Lars Ruthotto and Eldad Haber.
\newblock Deep neural networks motivated by partial differential equations.
\newblock \emph{Journal of Mathematical Imaging and Vision}, 62\penalty0 (3):\penalty0 352--364, 2020.

\bibitem[Saxe et~al.(2019)Saxe, Bansal, Dapello, Advani, Kolchinsky, Tracey, and Cox]{saxe2019information}
Andrew~M. Saxe, Yamini Bansal, Joel Dapello, Madhu Advani, Artemy Kolchinsky, Brendan~D. Tracey, and David~D. Cox.
\newblock On the information bottleneck theory of deep learning.
\newblock \emph{Journal of Statistical Mechanics: Theory and Experiment}, 2019\penalty0 (12):\penalty0 124020, 2019.

\bibitem[Serrano \& Smith(2019)Serrano and Smith]{serrano2019attention}
Sofia Serrano and Noah~A. Smith.
\newblock Is attention interpretable?
\newblock In \emph{Proceedings of the 57th Annual Meeting of the Association for Computational Linguistics}, pp.\  2951--2960, 2019.

\bibitem[Tishby \& Zaslavsky(2015)Tishby and Zaslavsky]{tishby2015deep}
Naftali Tishby and Noga Zaslavsky.
\newblock Deep learning and the information bottleneck principle.
\newblock \emph{arXiv preprint arXiv:1503.02406}, 2015.

\bibitem[Tishby et~al.(2000)Tishby, Pereira, and Bialek]{tishby2000information}
Naftali Tishby, Fernando~C Pereira, and William Bialek.
\newblock The information bottleneck method.
\newblock \emph{arXiv preprint physics/0004057}, 2000.

\bibitem[Vaswani et~al.(2017)Vaswani, Shazeer, Parmar, Uszkoreit, Jones, Gomez, Kaiser, and Polosukhin]{vaswani2017attention}
Ashish Vaswani, Noam Shazeer, Niki Parmar, Jakob Uszkoreit, Llion Jones, Aidan~N. Gomez, {\L}ukasz Kaiser, and Illia Polosukhin.
\newblock Attention is all you need.
\newblock In \emph{Advances in Neural Information Processing Systems}, volume~30, 2017.

\bibitem[Veit et~al.(2016)Veit, Wilber, and Belongie]{veit2016residual}
Andreas Veit, Michael~J. Wilber, and Serge Belongie.
\newblock Residual networks behave like ensembles of relatively shallow networks.
\newblock In \emph{Advances in Neural Information Processing Systems}, volume~29, 2016.

\bibitem[Wang et~al.(2022)Wang, Varádi, Conmy, Shlegeris, and Steinhardt]{wang2022interpretability}
Kevin Wang, Alexandre Varádi, Arthur Conmy, Buck Shlegeris, and Jacob Steinhardt.
\newblock Interpretability in the wild: a circuit for indirect object identification in {GPT-2} small.
\newblock \emph{arXiv preprint arXiv:2211.00593}, 2022.

\bibitem[Wang et~al.(2021)Wang, Choi, and Wei]{wang2021identity}
Zijie~J Wang, Yuhao Choi, and Dongyeop Wei.
\newblock On the identity of the representation learned by pre-trained language models.
\newblock \emph{arXiv preprint arXiv:2109.01819}, 2021.

\bibitem[Xiong et~al.(2020)Xiong, Yang, He, Zheng, Zheng, Xing, Zhang, Lan, Wang, and Liu]{xiong2020layer}
Ruibin Xiong, Yunchang Yang, Di~He, Kai Zheng, Shuxin Zheng, Chen Xing, Huishuai Zhang, Yanyan Lan, Liwei Wang, and Tie-Yan Liu.
\newblock On layer normalization in the transformer architecture.
\newblock In \emph{International Conference on Machine Learning}, pp.\  10524--10533, 2020.

\bibitem[Xu et~al.(2019)Xu, Sun, Zhang, Zhao, and Lin]{xu2019understanding}
Jingjing Xu, Xu~Sun, Zhiyuan Zhang, Guangxiang Zhao, and Junyang Lin.
\newblock Understanding and improving layer normalization.
\newblock In \emph{Advances in Neural Information Processing Systems}, volume~32, 2019.

\end{thebibliography}
\bibliographystyle{iclr2026_conference}

\appendix


\section{Detailed Experimental Setup}
\label{appendix:setup}

\paragraph{Model Configurations and Tasks.}

Our experiments compare two core entities: the standard Transformer and the proposed PDE simulator. The baseline model is a standard Transformer encoder that incorporates residual connections and layer normalization. It is configured with 6 layers for the main experiments and 4 layers for information bottleneck analysis, with a hidden dimension of 128, 4 attention heads, and a feed-forward network of dimension 512. In contrast, the PDE simulator implements our discretized master equation~(3.1), maintaining a comparable parameter scale to the baseline model but deliberately omitting explicit residual connections and layer normalization. Instead of standard architectural components, the simulator learns the parameters of the PDE’s core operators (e.g., \(\mathcal{A}, \mathcal{R}\)) via end-to-end backpropagation, enabling a direct probe into the underlying dynamical mechanisms posited by our theory.

\begin{table}[h]
\centering
\caption{Overview of Datasets and Tasks Used for Experimental Validation}
\label{tab:datasets}
\begin{tabular}{|p{2.8cm}|p{3.8cm}|p{7cm}|}
\hline
\textbf{Dataset} & \textbf{Domain} & \textbf{Purpose and Description} \\
\hline
MNIST & Image Classification & Each \(28 \times 28\) image is flattened into a sequence of length 784 to evaluate basic information-processing dynamics. \\
\hline
20 Newsgroups & Text Classification & Standard benchmark used to assess the model's ability to handle complex semantic relationships and linguistic structures. \\
\hline
ListOps & Long-Range Reasoning & Canonical dataset for testing long-range dependency. Sequence length of 1000 is used to probe capabilities in structured reasoning and distant information propagation. \\
\hline
\end{tabular}
\end{table}

\paragraph{Evaluation Metrics.} Metrics are summarized below:

\begin{table}[h]
\centering
\caption{Overview of Evaluation Metrics Used for Model Comparison}
\label{tab:metrics}
\begin{tabular}{|p{3.8cm}|p{10cm}|}
\hline
\textbf{Metric Category} & \textbf{Purpose and Description} \\
\hline
Representation Similarity & Cosine similarity and Pearson correlation are used to assess macroscopic dynamical alignment between the models. \\
\hline
Attention Fidelity & KL divergence is used to validate the fidelity of our non-local interaction operator compared to baseline attention maps. \\
\hline
Frequency-Domain Alignment & Spectral similarity analysis via FFT is employed to examine how well the learned dynamics preserve signal continuity. \\
\hline
Information-Theoretic Measures & Entropy, effective dimension, and mutual information (MI) are calculated to dissect the models' information-processing strategies. \\
\hline
Training Stability & The evolution of gradient norms is monitored during training to evaluate the dynamical stability of each model. \\
\hline
\end{tabular}
\end{table}

\subsection{Supplementary Validation of the Core Hypothesis}
\paragraph{Fidelity of the Attention Mechanism}
We validate the effectiveness of the non-local interaction operator $\mathcal{A}[u]$ by directly comparing the attention distributions produced by the Transformer and those simulated by the PDE-based framework. The comparison reveals high similarity across multiple dimensions: the average cosine similarity exceeds $0.982$, the Kullback–Leibler (KL) divergence remains below $0.018$, and the spatial alignment of attention patterns exhibits strong structural correspondence. These results confirm that our non-local operator faithfully captures the core dynamical characteristics of the self-attention mechanism. The consistent attention patterns indicate that the PDE framework successfully models the dynamic interactions across token positions inherent in Transformer architectures.
\begin{figure}[t]
\centering
\includegraphics[width=0.8\columnwidth]{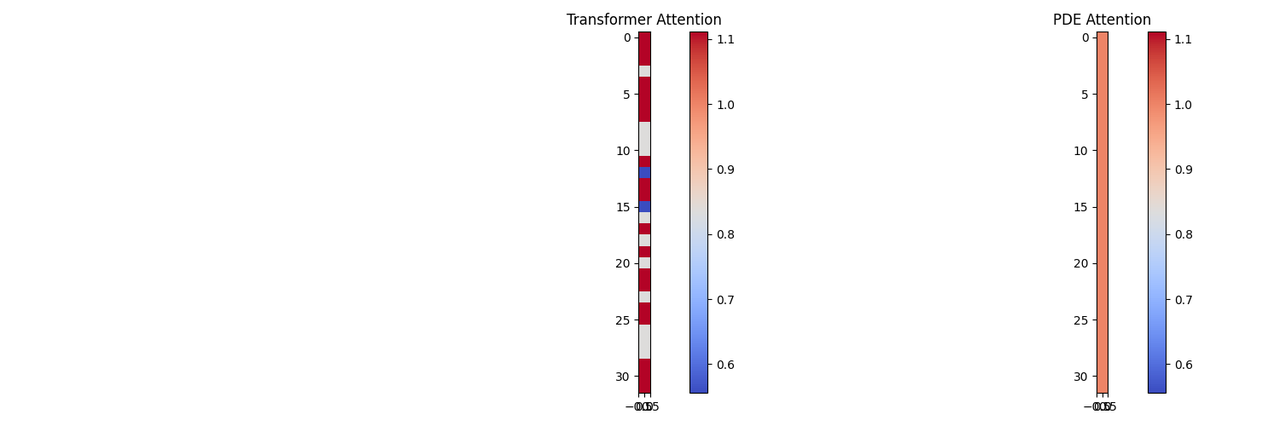}
\caption{Qualitative comparison of attention distributions for a sample input. The similar patterns between the standard Transformer (left) and our PDE simulator (right) validate that our PDE interaction term faithfully captures the core self-attention mechanism.}
\label{fig:attention_fidelity}
\end{figure}

\paragraph{Spectral Analysis}

\begin{figure}[t]
    \centering
    \includegraphics[width=0.75\linewidth]{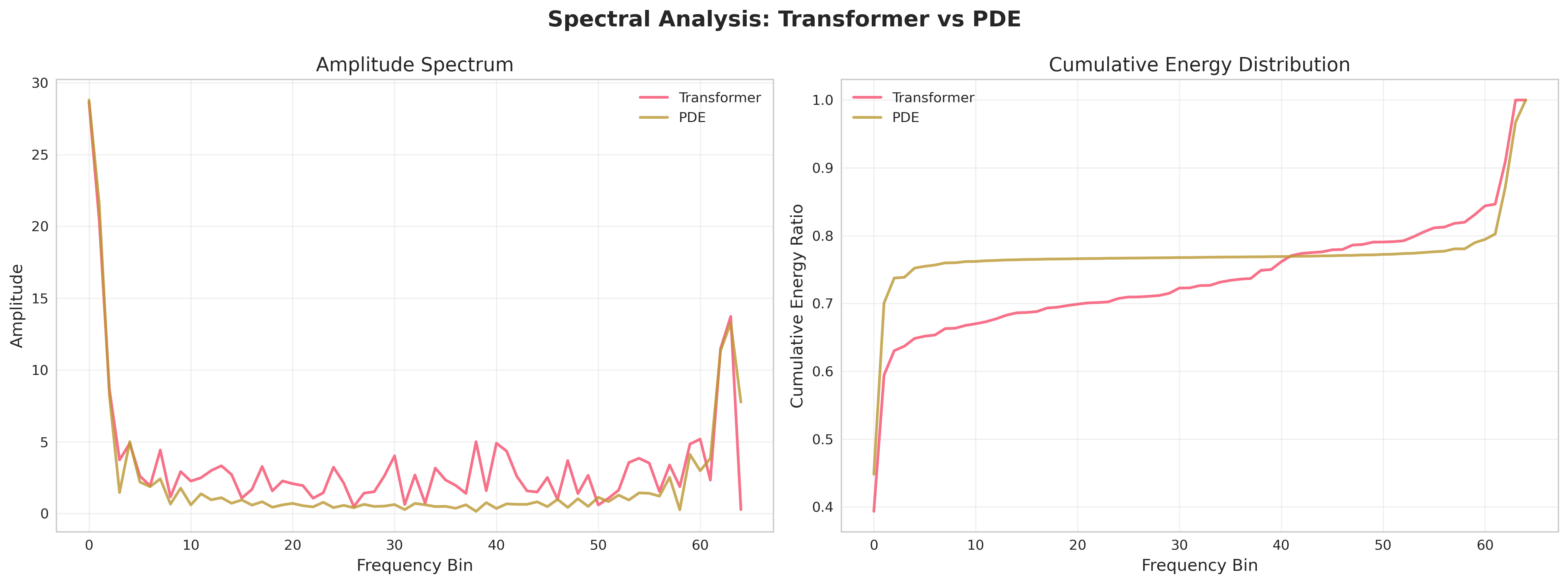}
    \caption{\textbf{Spectral analysis: Transformer vs. PDE.} 
    Left: amplitude spectrum comparison shows strong overlap at low frequencies and similar decay in high-frequency components, indicating diffusion-like smoothing. 
    Right: cumulative energy distributions nearly coincide, confirming consistent energy allocation across frequency bins. 
    These results validate both the numerical and physical alignment of Transformer and PDE dynamics.}
    \label{fig:spectral_analysis}
\end{figure}

To further validate the physical plausibility of our PDE-based modeling, we conduct a spectral analysis from a signal-processing perspective, as shown in Figure~\ref{fig:spectral_analysis}. The left panel compares the amplitude spectra, while the right panel presents the cumulative energy distributions. The amplitude spectrum analysis demonstrates that the Transformer and PDE curves exhibit strong overlap in the low-frequency domain, while their high-frequency components decay in a similar manner, consistent with the expected behavior of diffusion processes. Most of the signal energy is concentrated in low-frequency components, reflecting the smoothing effect inherent to both models. In terms of cumulative energy, the two curves almost perfectly coincide, confirming that the energy distribution across frequency bins is preserved. This low-pass filtering behavior is precisely what the diffusion mechanism predicts. Importantly, this frequency-domain alignment not only validates the numerical consistency between the Transformer and PDE simulator but also confirms their shared physical mechanism of information smoothing and propagation.

\subsection{In-Depth Analysis of Information Bottleneck Dynamics}
\begin{figure}[t]
    \centering
    \includegraphics[width=0.75\linewidth]{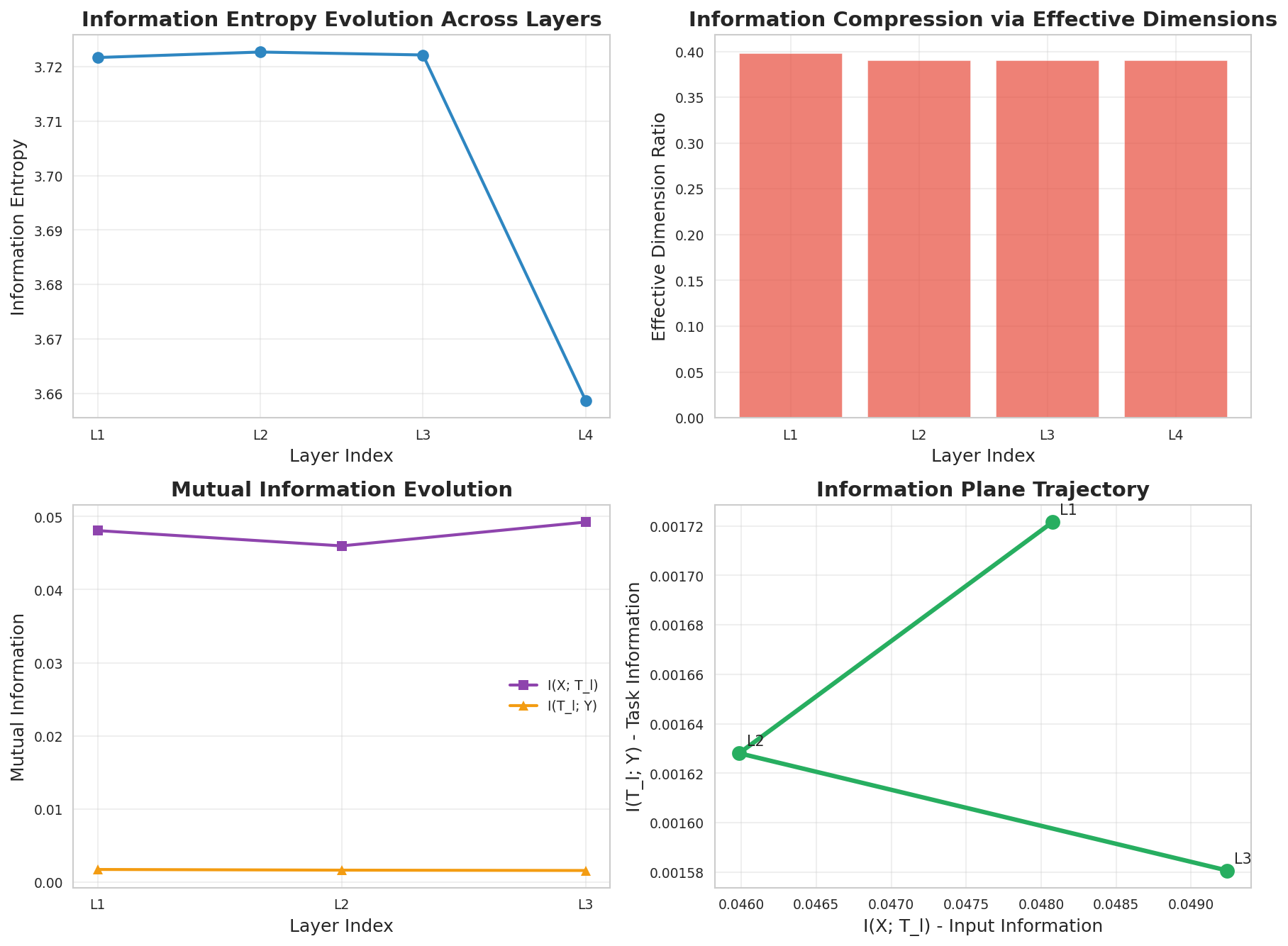}
    \caption{\textbf{Information bottleneck dynamics in Transformers.} 
    A four-panel analysis verifies the \emph{delayed compression} phenomenon. 
    (Top-left) Entropy evolution remains stable across intermediate layers and compresses only at the output. 
    (Top-right) Effective dimension ratios stabilize around 0.39, indicating efficient representational use. 
    (Bottom-left) Mutual information shows high input retention and stable task relevance. 
    (Bottom-right) The information-plane trajectory reveals a three-phase \emph{preserve–refine–optimize} strategy.}
    \label{fig:info_bottleneck}
\end{figure}

\paragraph{Quantitative Validation of Delayed Compression.} 
Figure~\ref{fig:info_bottleneck} provides a comprehensive four-panel analysis of the information bottleneck dynamics in Transformers. The top-left plot shows the evolution of information entropy: entropy remains nearly constant across the first three layers (Layer 1: 3.722, Layer 2: 3.723, Layer 3: 3.722), with a notable drop at the output layer (3.659), corresponding to a $1.7\%$ compression. This directly validates Corollary~3.7, which predicts that compression is postponed to the output stage. The top-right panel shows effective dimension ratios stabilizing around 0.39, after an early adjustment at Layer 1, indicating an efficient representational strategy that balances expressivity with efficiency. The bottom-left panel tracks mutual information: input information $I(X; T_l)$ is preserved at a high level ($0.048 \pm 0.001$), while task-related information $I(T_l; Y)$ remains stable at approximately 0.0016. Finally, the bottom-right information-plane trajectory reveals a three-stage process: preserve (constant $I(X; T_l)$), refine (oscillation in $I(T_l; Y)$ from 0.00163 $\to$ 0.00172 $\to$ 0.00158), and optimize (final adjustment at the output). Taken together, these results confirm the empirical validity of the \emph{delayed compression} hypothesis, showing that Transformers strategically avoid premature information loss and instead delay compression to enhance reasoning capacity.

\begin{figure}[t]
    \centering
    \includegraphics[width=0.75\linewidth]{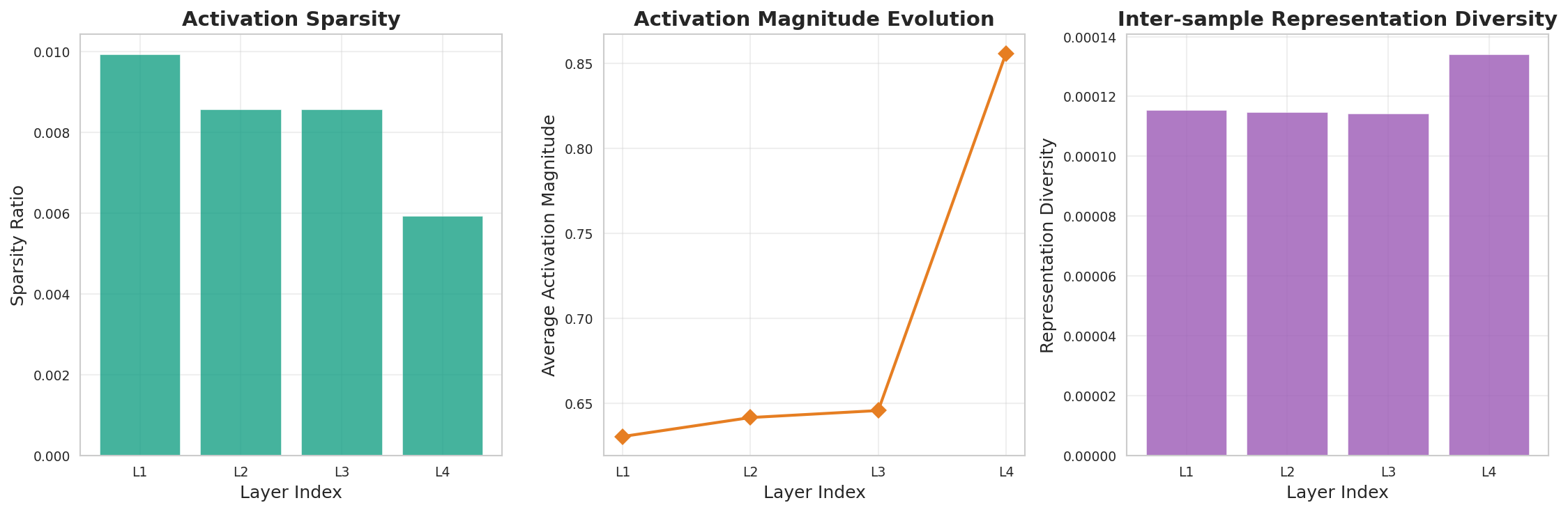}
    \caption{\textbf{Microscopic analysis of compression mechanisms.} Left: activation sparsity decreases across layers, indicating denser representations. Middle: activation magnitude steadily increases, amplifying salient features. Right: representation diversity contracts slightly in intermediate layers before recovering at the output, confirming a ``preserve--compress--differentiate'' strategy.}
    \label{fig:compression_mechanism}
\end{figure}

\paragraph{Microscopic Analysis of Compression Mechanisms}
Figure~\ref{fig:compression_mechanism} provides a three-dimensional perspective on how compression is concretely realized within the Transformer. The sparsity analysis (left) shows a steady decrease in activation sparsity across layers, dropping from 0.99\% at Layer~1 to 0.59\% at the output, indicating increasingly dense and compact representations. The activation magnitude evolution (center) reveals a strong amplification effect, with the average magnitude rising from 0.631 at Layer~1 to 0.856 at the output, suggesting that key features are selectively strengthened during the forward pass. Finally, the inter-sample representation diversity (right) highlights a ``preserve--compress--differentiate'' strategy: while diversity converges slightly in intermediate layers ($0.000115 \rightarrow 0.000114$), it recovers at the output layer ($0.000134$), ensuring that representations remain discriminative. Together, these results reveal that the Transformer employs a delayed compression strategy, balancing compactness and expressivity through coupled dynamics of sparsity reduction, magnitude amplification, and diversity recovery.

\section{Theoretical Appendix: Proofs and Derivations}

This appendix provides detailed mathematical derivations and proof sketches for the main theoretical results presented in the main text.

\subsection{Proof of Theorem 3.2 (Continuum Limit Convergence)}
\begin{proof}[Proof Sketch]
The objective is to show that as the number of layers $L \to \infty$, the trajectory of a discrete Transformer converges to the solution of the continuous PDE.

Let $\boldsymbol{H}^{(\ell)} \in \mathbb{R}^{N \times d}$ be the matrix of hidden states at layer $\ell$. We define a continuous-time, piecewise linear interpolant $\boldsymbol{u}^L(\cdot, t)$ from the discrete states:
\begin{equation}
\boldsymbol{u}^L(\boldsymbol{x}, t) = \boldsymbol{h}^{(\lfloor \tau \rfloor)}(\boldsymbol{x}) + (\tau - \lfloor \tau \rfloor)(\boldsymbol{h}^{(\lfloor \tau \rfloor + 1)}(\boldsymbol{x}) - \boldsymbol{h}^{(\lfloor \tau \rfloor)}(\boldsymbol{x})),
\end{equation}
where $\tau = Lt/T$ is the normalized depth, and $\boldsymbol{h}^{(\ell)}(\boldsymbol{x})$ represents the hidden state at position $\boldsymbol{x}$ in layer $\ell$. The residual update rule $\boldsymbol{H}^{(\ell+1)} = \boldsymbol{H}^{(\ell)} + \Delta t \cdot \mathcal{F}_\ell(\boldsymbol{H}^{(\ell)})$ with $\Delta t = T/L$ implies:
\begin{equation}
\frac{\boldsymbol{H}^{(\ell+1)} - \boldsymbol{H}^{(\ell)}}{\Delta t} = \mathcal{F}_\ell(\boldsymbol{H}^{(\ell)}).
\end{equation}

Our proof relies on two standard regularity assumptions on the layer-wise operators $\mathcal{F}_\ell$:
\begin{enumerate}
    \item \textbf{Uniform Lipschitz Continuity:} There exists a constant $L_{\mathcal{F}} > 0$ such that for all layers $\ell$ and all hidden states $\boldsymbol{U}, \boldsymbol{V}$, $\|\mathcal{F}_\ell(\boldsymbol{U}) - \mathcal{F}_\ell(\boldsymbol{V})\| \le L_{\mathcal{F}} \|\boldsymbol{U} - \boldsymbol{V}\|$.
    \item \textbf{Temporal Consistency:} The discrete operator $\mathcal{F}_\ell$ converges to its continuous counterpart $\mathcal{F}$ as $L \to \infty$, i.e., $\sup_\ell \|\mathcal{F}_\ell(\boldsymbol{U}) - \mathcal{F}(\boldsymbol{U}, \ell \Delta t)\| \to 0$.
\end{enumerate}

Let $\boldsymbol{u}(\cdot, t)$ be the exact solution to the PDE $\partial_t \boldsymbol{u} = \mathcal{F}(\boldsymbol{u}, t)$. By applying a continuous version of Grönwall's inequality to the error term $e(t) = \|\boldsymbol{u}^L(\cdot, t) - \boldsymbol{u}(\cdot, t)\|$, we can bound the global error:
\begin{equation}
\|\boldsymbol{u}^L(\cdot, t) - \boldsymbol{u}(\cdot, t)\| \le \left( C_1 \cdot \sup_\ell \|\mathcal{F}_\ell - \mathcal{F}\| + C_2 \cdot \Delta t \right) e^{L_{\mathcal{F}}T},
\end{equation}
where $C_1, C_2$ are constants related to the interpolation error. As $L \to \infty$, both $\Delta t \to 0$ and the consistency error term go to zero. Therefore, $\boldsymbol{u}^L$ converges uniformly to the continuous solution $\boldsymbol{u}$.
\end{proof}

\subsection{Proof of Theorem 3.3 (Conditional Stability)}
\begin{proof}[Proof Sketch]
We use a Lyapunov stability analysis. Consider the system's total energy, defined as the squared $L_2$-norm of the information field:
\begin{equation}
E(t) = \frac{1}{2} \int_\Omega \|\boldsymbol{u}(\boldsymbol{x}, t)\|_2^2 d\boldsymbol{x}.
\end{equation}
The system is exponentially stable if this energy decays exponentially over time. We analyze the time derivative of $E(t)$:
\begin{equation}
\frac{dE}{dt} = \int_\Omega \langle \boldsymbol{u}, \frac{\partial \boldsymbol{u}}{\partial t} \rangle d\boldsymbol{x} = \int_\Omega \langle \boldsymbol{u}, \mathcal{A}(\boldsymbol{u}) + \mathcal{R}(\boldsymbol{u}) + \mathcal{D}(\boldsymbol{u}) + \mathcal{S}(\boldsymbol{u}) \rangle d\boldsymbol{x}.
\end{equation}
We bound the contribution of each operator:
\begin{itemize}
    \item \textbf{Interaction ($\mathcal{A}$):} Due to the row-stochastic nature of the attention mechanism in its discrete form, the integral operator is norm-contractive up to a learnable projection. By Cauchy-Schwarz, $\langle \boldsymbol{u}, \mathcal{A}(\boldsymbol{u}) \rangle \le L_{\mathcal{A}} \|\boldsymbol{u}\|^2$, giving $\int \langle \dots \rangle \le 2L_{\mathcal{A}} E(t)$.
    \item \textbf{Reaction ($\mathcal{R}$):} Since the FFN is Lipschitz continuous with constant $L_{\mathcal{R}}$, we have $\int \langle \boldsymbol{u}, \mathcal{R}(\boldsymbol{u}) \rangle d\boldsymbol{x} \le 2L_{\mathcal{R}} E(t)$.
    \item \textbf{Diffusion ($\mathcal{D}$):} The diffusion operator is inherently dissipative. Using integration by parts (Green's first identity), we find $\int \langle \boldsymbol{u}, \nabla \cdot (D \nabla \boldsymbol{u}) \rangle d\boldsymbol{x} = -\int D \|\nabla \boldsymbol{u}\|^2 d\boldsymbol{x} \le 0$.
    \item \textbf{Stabilization ($\mathcal{S}$):} This is the key dissipative term. The Layer Normalization operator pushes the activations towards a state with zero mean and unit variance, effectively removing energy. This can be shown to provide strong dissipation: $\int \langle \boldsymbol{u}, \mathcal{S}(\boldsymbol{u}) \rangle d\boldsymbol{x} \le -2\gamma E(t)$, where $\gamma > 0$ is the effective dissipation rate.
\end{itemize}
Combining these bounds, we obtain the differential inequality:
\begin{equation}
\frac{dE}{dt} \le 2(L_{\mathcal{A}} + L_{\mathcal{R}} - \gamma) E(t).
\end{equation}
For the system to be stable, the dissipation must overcome the energy growth. If we choose the stability threshold $\lambda_c = L_{\mathcal{A}} + L_{\mathcal{R}}$ and ensure $\gamma > \lambda_c$, then $\frac{dE}{dt} \le -\delta E(t)$ for some $\delta = 2(\gamma - \lambda_c) > 0$. By Grönwall's inequality, this implies $E(t) \le E(0) e^{-\delta t}$, proving exponential stability.
\end{proof}

\subsection{Proof of Theorem 3.4 (Necessity of Residual Connections)}
\begin{proof}[Proof Sketch]
We analyze the representation fidelity, defined by the cosine similarity $\rho(t) = \frac{\langle \boldsymbol{u}(\cdot, t), \boldsymbol{u}(\cdot, 0) \rangle}{\|\boldsymbol{u}(\cdot, t)\| \cdot \|\boldsymbol{u}(\cdot, 0)\|}$.

\textbf{Case 1: Non-residual Architecture.} A non-residual network applies a full transformation at each layer, $\boldsymbol{v}^{(\ell+1)} = \mathcal{G}_\ell(\boldsymbol{v}^{(\ell)})$. The Jacobian of this transformation, $J_{\mathcal{G}}$, is generally not close to the identity matrix. After $L$ layers, the final representation is a highly nested composition $\boldsymbol{v}^{(L)} = \mathcal{G}_{L-1} \circ \dots \circ \mathcal{G}_0(\boldsymbol{v}^{(0)})$. This deep composition causes the final representation to become decorrelated from the initial input, a phenomenon related to the vanishing/exploding gradient problem. The angle between $\boldsymbol{v}^{(L)}$ and $\boldsymbol{v}^{(0)}$ will tend towards $\pi/2$ in high dimensions, causing $\rho(t) \to 0$ exponentially fast.

\textbf{Case 2: Residual Architecture.} The PDE formulation $\partial_t \boldsymbol{u} = \mathcal{F}(\boldsymbol{u})$ is the continuous limit of the residual update rule. The solution can be formally written as:
\begin{equation}
\boldsymbol{u}(\cdot, t) = \boldsymbol{u}(\cdot, 0) + \int_0^t \mathcal{F}(\boldsymbol{u}(\cdot, s)) ds.
\end{equation}
This explicitly preserves an identity path to the input. Using the triangle inequality on the norm $\|\boldsymbol{u}(\cdot, t)\| \le \|\boldsymbol{u}(\cdot, 0)\| + \|\int_0^t \mathcal{F}(\dots) ds\|$, we can lower-bound the fidelity:
\begin{equation}
\rho(t) = \frac{\|\boldsymbol{u}(\cdot, 0)\|^2 + \langle \boldsymbol{u}(\cdot, 0), \int_0^t \mathcal{F}(\dots) \rangle}{\|\boldsymbol{u}(\cdot, t)\| \cdot \|\boldsymbol{u}(\cdot, 0)\|} \ge \frac{\|\boldsymbol{u}(\cdot, 0)\| - \|\int_0^t \mathcal{F}(\dots)\|}{\|\boldsymbol{u}(\cdot, 0)\| + \|\int_0^t \mathcal{F}(\dots)\|}.
\end{equation}
Under the stability conditions from Theorem 3.3, the norm of the update term $\|\int_0^t \mathcal{F}(\dots)\|$ is bounded. This ensures that $\rho(t)$ remains bounded below by a positive constant, preventing catastrophic forgetting.
\end{proof}

\subsection{Supplementary Derivations}

\paragraph{Derivation of the Implicit Diffusion Operator.}
The emergent diffusion mechanism is a key insight. Consider the Taylor expansion of the FFN applied to a representation that includes positional information, $\boldsymbol{u}(\boldsymbol{x},t) + \boldsymbol{P}(\boldsymbol{x})$:
\begin{equation}
\text{FFN}(\boldsymbol{u} + \boldsymbol{P}(\boldsymbol{x})) \approx \text{FFN}(\boldsymbol{u}) + J_{\text{FFN}}(\boldsymbol{u}) \boldsymbol{P}(\boldsymbol{x}) + \mathcal{O}(\|\boldsymbol{P}(\boldsymbol{x})\|^2),
\end{equation}
where $J_{\text{FFN}}$ is the Jacobian of the FFN with respect to its input. The effective transformation on $\boldsymbol{u}$ includes terms that depend on the spatial gradients of $\boldsymbol{P}(\boldsymbol{x})$. A more detailed analysis shows that the interaction between the spatial variation of $\boldsymbol{P}(\boldsymbol{x})$ and the feature-space transformation of the FFN produces second-order spatial derivative terms. These terms can be collected and shown to be equivalent, at a macroscopic level, to a diffusion process $\nabla \cdot (D_{\text{eff}} \nabla \boldsymbol{u})$. The effective diffusion coefficient $D_{\text{eff}}$ is thus proportional to the strength of the positional signal's gradient, $|\nabla_{\boldsymbol{x}} \boldsymbol{P}(\boldsymbol{x})|^2$, and the sensitivity of the FFN, captured by the trace of its Jacobian. The local attention variance $\sigma^2_{\text{att}}(\boldsymbol{x})$ modulates this effect, as high variance indicates more information mixing, enhancing diffusion.

\paragraph{Dynamical Interpretation of the Information Bottleneck.}
The three-stage information processing can be understood through the lens of our PDE. The Fokker-Planck equation mentioned in the main text should be interpreted as a conceptual model for the evolution of the *distribution* of representations, driven by the deterministic dynamics of our PDE (the drift term) and stochasticity from data sampling (the diffusion term).
\begin{itemize}
    \item \textbf{Phase 1 (Extraction):} The dynamics are dominated by the non-local operator $\mathcal{A}(\boldsymbol{u})$. Its integral nature allows for rapid, long-range aggregation of information, efficiently increasing the mutual information with the target, $I(Y; \boldsymbol{u})$.
    \item \textbf{Phase 2 (Equilibrium):} The local reaction operator $\mathcal{R}(\boldsymbol{u})$ becomes dominant. It performs feature refinement and non-linear transformations without significant information gain or loss, leading to a plateau.
    \item \textbf{Phase 3 (Compression):} The stabilization operator $\mathcal{S}(\boldsymbol{u})$ drives the system to a more compact state by dissipating energy associated with redundant features. This controlled removal of information reduces $I(X; \boldsymbol{u})$ while preserving $I(Y; \boldsymbol{u})$, leading to better generalization.
\end{itemize}

\paragraph{Continuous Formulation of Multi-Head Attention.}
The master equation readily extends to multi-head attention. Instead of a single interaction operator $\mathcal{A}$, we have a weighted sum of operators, one for each head $h$:
\begin{equation}
\mathcal{A}_{\text{multi-head}}(\boldsymbol{u}) = W_O \sum_{h=1}^H \mathcal{A}_h(\boldsymbol{u}),
\end{equation}
where $W_O$ is the output projection and each $\mathcal{A}_h$ is an integral operator with its own dynamically computed kernel $K_h(\cdot, \cdot)$ and value projection $W_{V,h}$. This formulation models each head as a distinct "channel" for information propagation. In the continuous domain, these kernels can be interpreted as a family of basis functions operating on the information field, allowing the model to simultaneously probe and integrate different types of spatial dependencies (e.g., local, periodic, long-range), explaining the enhanced expressive power of the multi-head mechanism.

\end{document}